\documentclass{ecai}
\pdfoutput=1
\usepackage{times}
\usepackage{graphicx}
\usepackage{latexsym}
\usepackage{hyperref}
\usepackage{cleveref}
\usepackage{algorithmic}
\usepackage[linesnumbered,ruled,noend]{algorithm2e}
\usepackage{booktabs}
\usepackage{textcomp}
\usepackage{subfigure}
\usepackage{stfloats}
\usepackage{amsmath,amssymb,amsfonts}
\usepackage{color}
\usepackage{epstopdf}
\usepackage{bm}
\usepackage{mdwlist}

\begin{document}

\title{DeepGS: Deep Representation Learning of Graphs and Sequences for Drug-Target Binding Affinity Prediction }

\author{Xuan Lin\institute{College of Computer Science and Technology, Hunan University, China}
}

\maketitle
\bibliographystyle{ecai}

\begin{abstract}
Accurately predicting drug-target binding affinity (DTA) in silico is a key task in drug discovery. Most of the conventional DTA prediction methods are simulation-based, which rely heavily on domain knowledge or the assumption of having the 3D structure of the targets, which are often difficult to obtain. Meanwhile, traditional machine learning-based methods apply various features and descriptors, and simply depend on the similarities between drug-target pairs. Recently, with the increasing amount of affinity data available and the success of deep representation learning models on various domains, deep learning techniques have been applied to DTA prediction. However, these methods consider either label/one-hot encodings or the topological structure of molecules, without considering the local chemical context of amino acids and SMILES sequences. Motivated by this, we propose a novel end-to-end learning framework, called \textsf{\footnotesize DeepGS}, which uses deep neural networks to extract  the local chemical context from amino acids and SMILES sequences, as well as the molecular structure from the drugs. To assist the operations   on the symbolic data, we propose to use advanced embedding techniques (i.e., {Smi2Vec} and {Prot2Vec}) to encode the   amino acids and SMILES sequences to a distributed representation. Meanwhile, we suggest a new molecular structure modeling approach that works well under our framework. We have conducted extensive experiments to compare our proposed method with state-of-the-art models including {KronRLS}, {SimBoost}, {DeepDTA} and {DeepCPI}. Extensive experimental results demonstrate the superiorities and competitiveness of {DeepGS}.
\end{abstract}

\section{Introduction}\label{sec:introduction}
Effectively predicting drug-target binding affinity (DTA) is one of the important problems in drug discovery. Drugs (or ligands) \cite{AAAI2018li} are chemical compounds, each of which can be represented by both a molecule graph with atoms as nodes and chemical bonds as edges, and a string obtained from the Simplified Molecular Input Line Entry System (SMILES)~\cite{SMILES1988Weininger}. Targets (or proteins) are sequences of amino acids. Binding affinity indicates the strength of the interactions of drug-target pairs. Through binding, drugs can have a positive or negative influence on functions carried out by proteins, affecting the disease conditions \cite{you2018graph}.
By understanding drug-target binding affinity, it is possible to find out candidate drugs that are able to inhibit the target/protein and benefits many other bioinformatic applications \textcolor{black}{\cite{suarez2016using,xuanlin2019bib,BIBM2019quan}}. As a result, DTA prediction has received much attention in recent years \cite{campillos2008drug,keiser2009predicting,AAAI2017xiao}.

Early approaches for DTA prediction can be roughly classified into two types: (\romannumeral 1) simulation-based methods, and (\romannumeral 2) traditional machine learning-based methods.
Simulation-based methods rely on \textcolor{black}{domain} knowledge \cite{N9-kdd} or {the 3D structure of target/protein \cite{trott2010autodock,fout2017protein}}, which are often difficult to obtain. Meanwhile, traditional machine learning-based methods apply various features \textcolor{black}{\cite{van2011gaussian,N8-kdd}} and descriptors  \textcolor{black}{\cite{faulon2007genome,jacob2008protein,duvenaud2015convolutional}}, and simply depend on the similarities between drug-target pairs \textcolor{black}{\cite{wang2013prediction,N11-kdd,ma2018drug}}. Recently, owing to the remarkable success in various machine learning tasks (\emph{e.g.,} image recognition and natural language processing), deep learning-based methods are also exploited for DTA prediction \cite{ozturk2018deepdta}. These methods consider either label/one-hot encodings or the topological structure of molecules, they, however, do not consider the local chemical context of amino acids and SMILES sequences. It is easily understood that the topological structure information provides an overview of how the atoms are connected, while the local chemical context reveals the functionality of the atoms, like the semantic meaning of a word in a sentence. These two types of information complement each other and are both important for DTA prediction. 
It should be meaningful and interesting to take these two types of information consideration together.  To this end, this paper proposes a novel end-to-end learning framework for DTA prediction, namely \textbf{Deep} representation learning framework for \textbf{G}raphs and \textbf{S}equences (DeepGS).

In a nutshell, our framework consists of three major building blocks. One of the major blocks learns low-dimension vector representations for target/protein sequences, using a convolutional neural network (CNN). The other two blocks learn two representations for drugs, by using a graph attention network (GAT) and a bi-directional gate recurrent unit (BiGRU), respectively. Specifically, (\textbf{\romannumeral 1}) the CNN and BiGRU blocks extract local chemical context information of amino acids in targets and atoms in drugs, respectively. Since the label/one-hot encodings of amino acids and atoms often neglect the context information, and motivated by the idea of Word2Vec \cite{mikolov2013distributed}, we leverage advanced techniques, \textit{Smi2Vec} and \textit{Prot2Vec}, to encode the amino acids and atoms to a distributed representation, before plugging them to CNN and BiGRU. (\textbf{\romannumeral 2}) The newly designed GAT-based molecular structure modeling approach extracts the topological features of drugs, by aggregating the representations of $r$-radius subgraphs. (\textbf{\romannumeral 3}) The learned representations for both drugs and targets are then passed to a neural network to predict  the binding affinity.

Different from the existing simulation-based methods, our framework needs neither expert knowledge nor 3D structure of the targets, and so it could be more easy-to-use.  Additionally, the proposed framework takes advantage of the local chemical context information of atoms/amino acids in drugs/proteins and uses a newly designed molecular structure modeling approach, which differ DeepGS from the existing deep learning models.
To summarize, the main contributions of this paper are listed as follows:

\begin{itemize*}
\item We propose  a novel model {DeepGS} for DTA prediction. To the best of knowledge, this work is the first to 
consider  both local chemical context and topological structure   to learn the interaction between drugs and targets.
\item We conduct extensive experiments to study the performance of our proposed method, based on both small and large benchmarking   datasets. The experimental results demonstrate  (\romannumeral 1) the promising performance of our proposed model, (\romannumeral 2) considering jointly local chemical context and topological structure is effective, and (\romannumeral 3) the newly designed  molecular structure modeling approach works well under our proposed framework. (The codes of our method are available at https://github.com/jacklin18/DeepGS.)
\end{itemize*}

The rest of the paper is organized as follows. In Section \ref{sec:method}, we introduce the proposed method for drug-target binding affinity prediction. In Section \ref{sec:experiment}, we report and analyze the performance of our method. Section \ref{sec:relatedwork} reviews the related work. Finally, we conclude the paper in Section \ref{sec:conclusion}.

\section{The Proposed Method}\label{sec:method}
In this section, we first provide an overview of the proposed DeepGS framework (Section \ref{Overview}). Then, we introduce the representation learning for drugs and targets, respectively (Sections \ref{sec:drug}$\sim$\ref{sec:target}). Finally, we discuss the binding affinity prediction with DeepGS  (Section \ref{DTI prediction}).

\subsection{Overview of DeepGS}\label{Overview}

\begin{figure}[t]
\centerline{\includegraphics[width=1.0\linewidth]{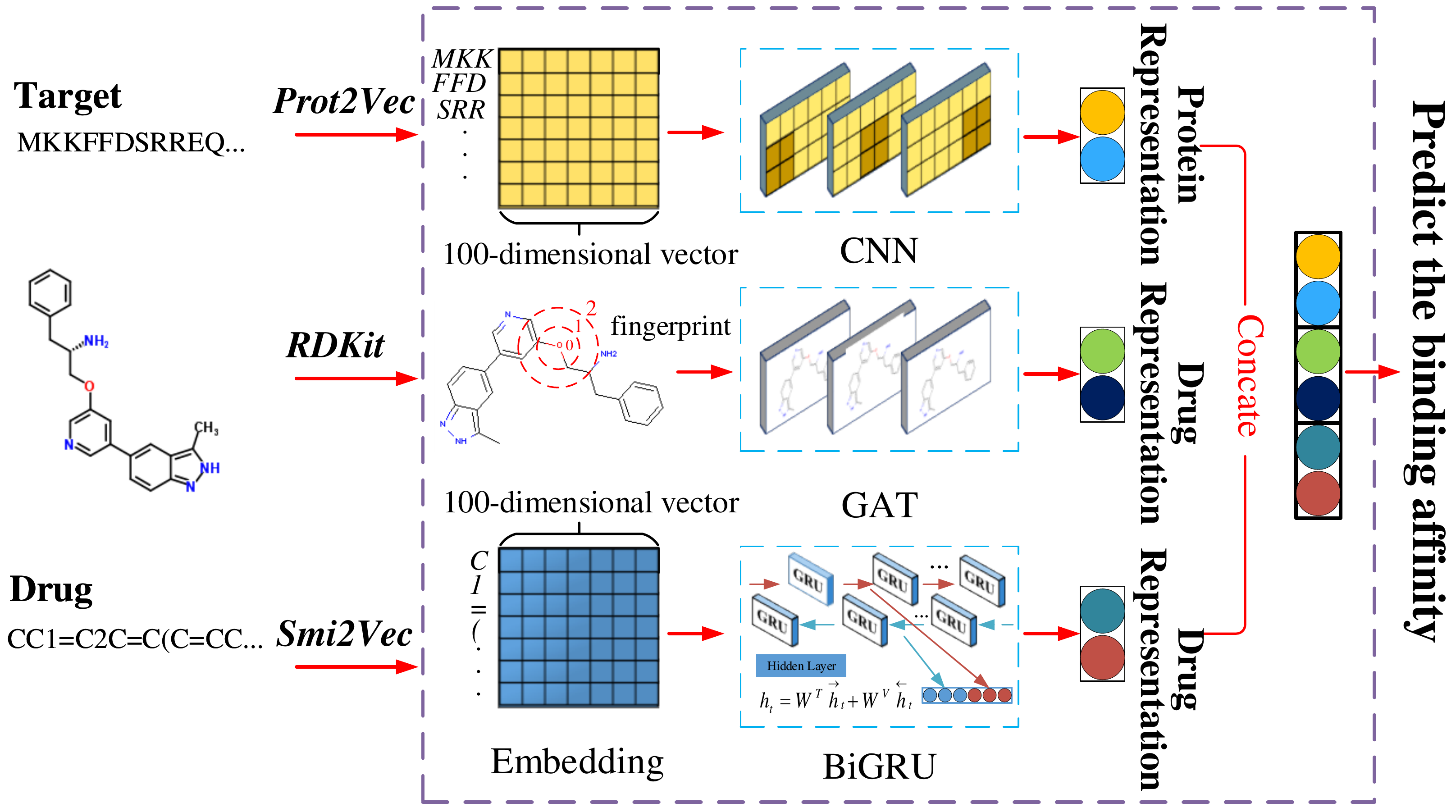}}
\caption{Overview of DeepGS.}
\label{overview}
\end{figure}

Figure \ref{overview} shows the overview of DeepGS. It takes the symbolic sequences of  target/protein and  drug, as well as the molecular structure of the drug as the input. It outputs the binding affinity for the drug-target pair.
Remind that the central idea of DeepGS is to consider both local chemical context and the molecular structure, by using  some embedding techniques (i.e., Smi2Vec and Prot2Vec) to encode the amino acids and atoms to a distributed representation.
 Therefore, we design DeepGS as a  three-step framework for DTA prediction:
\begin{enumerate*}
\item Encoding symbolic tokens in target/drug sequences;
\item Encoding the whole drug/target sequences and the molecular structure of the drug;
\item Predicting the binding affinity value based on the encodings of the drug and the target.
\end{enumerate*}

Specifically, motivated by   Word2Vec \cite{mikolov2013distributed}, in \textbf{\small the first step} we encode the symbols in the sequence of the target/protein and the drug to a distributed representation, by using \textit{Prot2Vec} and \textit{Smi2Vec}, respectively. Then, the sequences can be transformed into matrices, where each row is the representation of a symbol in the sequences. In \textbf{\small the second step} we extract features, from the drug/target matrices and the molecule graph, to encode the whole sequences and graph. For the target/protein, we consider the local chemical context of the amino acids, by using a convolutional neural network (CNN). For the drug, we consider both the molecular structure and the local chemical context. Particularly, since the molecular structure can be represented by a graph,   we suggest a graph attention network (GAT) based approach to extract the topological information of the drug. In the meantime, the local chemical context of atoms in the drug is captured, by using a bi-directional gated recurrent unit (BiGRU). As a result, we obtain a latent representation for the target and two latent presentations for the drug. To predict the binding affinity, \textbf{\small in the third step} DeepGS inputs the concatenation of the three latent representations to a stack of fully connected layers, and outputs a real value binding affinity. Next, we present the details of our method.

\subsection{Representation Learning for Drug}\label{sec:drug}
\subsubsection{Local Chemical Context Modeling}
Drugs are often presented in the format of SMILES (Simplified Molecular-Input Line-Entry System), a specification in the form of a line notation for describing the structure of chemical compound \cite{SMILES1988Weininger}. For example, the SMILES string of the drug in Figure \ref{overview} is ``CC1=C2C=C(C=CC...'', which is a sequence of atoms and covalent bonds. For ease of representation, we consider both atoms and covalent bonds as symbolic tokens, and so the SMILES sequence is a sequence of symbols. To encode the SMILES sequence, existing deep learning approaches such as \textit{DeepDTA} \cite{tsubaki2018compound} use label/one-hot encoding to represent each symbol in the SMILES sequence. However, label/one-hot encoding often neglects the context of the symbol, and thus cannot reveal the functionality of the symbol within the context. To remedy this, we propose to use \textit{Smi2Vec} \cite{BIBM2018Quan}, a method similar to Word2Vec \cite{mikolov2013distributed,ijcai2018-575,quan2019efficient}, to represent the symbols in the SMILES sequence. \textcolor{black}{Algorithm \ref{Smi2Vec} shows the pseudo-codes of encoding SMILES symbols, based on the pre-trained \textit{Smi2Vec} embeddings. In general, a SMILES string with fixed length,  say $m$, is divided into a separate atom or symbol (Line 1). Then, it \textit{maps} the atom by looking up each of the atom embeddings from the pre-trained dictionary, while it randomly generates values if it is not in the dictionary (Lines 2-6). Finally, it constructs an atom matrix $A$ by aggregating embedding vectors (Lines 7-8), where each line represents the pre-trained vector of an atom.}

\begin{algorithm}[t]
\caption{Smi2Vec}
\label{Smi2Vec}
\KwIn {a molecule $s$ in the format of SMILES, dictionary $D$, atom vector's fixed length \textit{m}, vector dimension $d$.}
\KwOut {atom matrix $A$}

atom set $\{x_j | 1\leq j<|s|\} \longleftarrow \textit{split}(s$)

\For {j=1 to m}
{
\If {$x_j \notin dictionary$} {embedding vector $a_j \longleftarrow$ randomly generated value $\in \Re^d$}

\Else {$a_j \xleftarrow{map} x_j$ // by using $D$}
}
atom matrix $A \longleftarrow \sum_{j=1}^{m} a_{j}$

\Return $A \in \Re^{m\times{d}}$
\end{algorithm}

Motivated by the gate function in GRU \cite{chung2015gated}, we apply a 1-layer BiGRU on the resulting matrix to obtain a latent representation of the drug, which allows us to   model the local chemical context. Note that BiGRU takes a fix-sized of matrix as the input, while the length of SMILES strings may vary. One simple solution is to fix the length of input sequence at approximately average length of the SMILES string in the dataset, and apply zero-paddings at the end of the input sequences. 
As we will show later in Section \ref{sec:experiment}, an appropriate length (e.g., larger than the average length of sequences in the dataset) does not make the performance of our framework   change a lot. Considering the training efficiency and the DTA performance, it is suggested that a small  number   is a good trade-off between efficiency and performance.

\subsubsection{Molecular Structure Modeling}\label{subsec:structureModel}
In addition to the local chemical context, we exploit the molecular structure to uncover how the atoms connect in the drug. The molecular structure is an important cue for DTA prediction \cite{tsubaki2018compound}.
To achieve this, we can first use   the RDKit \cite{landrum2006rdkit} tool to transform SMILES string of a chemical compound into a molecule graph $G=(V, E)$, in which  the node $v_i \in V$ represents the $i$-th atom, and $e_{ij} \in E$ represents the chemical bond between the $i$-th and the $j$-th atoms. Then, we  can learn a graph attention network (GAT) \cite{velivckovic2017graph} from the molecule graphs $G$.
To apply GAT on molecule graph, we can encode all atoms and chemical bonds to a $d$-dimensional vector, and aggregates the information from the $r$-radius subgraph for each atom in the molecular graph, where $r$ is the number of hops from an atom. \textcolor{black}{Algorithm \ref{GNN1} shows the pseudo-codes of applying \textcolor{black}{GAT} on molecule graphs. Specifically, it  first computes an initial vector concatenating the fingerprint (i.e., the $r$-radius subgraph) and the adjacent information for each atom (Lines 1-4). Here   it leverages   Weisfeiler-Lehman algorithm to extract the fingerprint of the atoms. Then, it updates the atom vectors by propagating the information from its neighboring nodes (Lines 5-6). Finally, it aggregates the atom vectors to obtain the representation of the molecule (Line 7), each of which contains the $r$-radius subgraph information.}

\begin{algorithm}[h]
\caption{\textcolor{black}{GAT} on molecule graph}
\label{GNN1}
\KwIn{Molecule graph $G = (V, E)$, radius $R$}
\KwOut{a vector $y_{molecule}$ for a molecule}

\For {each node $v_i \in V$} {
$adj(v_i) \leftarrow extract\_adjacency(G)$



$fp(v_i) \leftarrow extract\_fingerprints(v_i, G, R)$

$V_{in} \leftarrow [adj(v_i) ; fp(v_i)]$

}

\For{each node $v_i \in V$}{
      update $V_{in} \leftarrow V_{in} + \sum_{v_{j} \in Neighbors({v_{in}})} {\textit{GATConv}(V_{j})}$
       }

\Return $y_{molecule \leftarrow \sum_{v_1}^{V} V_{in}}$
\end{algorithm}
\vspace{-2ex}

\subsection{Representation Learning for Target/Protein}\label{sec:target}

Targets/proteins are often represented as a sequence of amino acids (\emph{e.g.}, MKKFFDSRREQ... shown in Figure \ref{overview}). Similar to the SMILES string, we propose to first encode the amino acids into a $d$-dimensional vector following \textit{Prot2Vec} \cite{asgari2015continuous}, which allows us to capture local chemical information in targets/proteins.  As a single amino acid often makes no sense, we apply a fixed-length $N$-gram splitting approach to partition the sequence into meaningful ``biological words''. \textcolor{black}{Note that,  here the sequence refers to  the \textit{fixed-length input protein sequence} (instead of   the full sequences), which is preprocessed as similar as we handle the SMILES strings (recall Section \ref{sec:drug}).} 
Compared to the commonly used label encoding methods, the fixed-length $N$-gram divides the sequence into a sequence of $N$-grams. Each $N$-gram is considered as a ``biological word''. Intuitively, it can generates more ``words context'' than label encoded by one-hot encoding.

Considering that there are generally 20 kinds of amino acids, rendering that the maximum number of possible N-grams is 20$^{N}$. To make trade-off between the training feasibility and vocabulary size, in our paper we define $N=3$. Specifically, given a protein or target sequence $L = \{x_{i} | (i=1, 2, ..., |l|)\}$, where $x_i$ represents the $i$-th amino acid and $|l|$ represents the sequence length, the fixed-length 3-gram splitting method partitions the sequence into the following 3-grams, each of which is a biological word consisting of 3 amino acids: $[x_{1}; x_{2}; x_{3}], [x_{4}; x_{5}; x_{6}],..., [x_{|l|-2}; x_{|l|-1}; x_{|l|}]$.
For each biological word, we map it to an embedding vector by looking up a pre-trained embedding dictionary for 9048 words \cite{asgari2015continuous}, which is obtained from Swiss-Prot (https://www.uniprot.org/) with 560,118 manually annotated sequences. As a result, we transform each target sequence to a matrix, in which each row is the embedding of a biological word. The matrix is then fed into a CNN to extract the local chemical context of the target.  It is worth noting that,  different from the early ligand-based approach \cite{jacob2008protein} that neglects the local context information in targets/proteins, our solution above leverages the embedding technique to learn {the representation} from the protein sequence.   

\subsection{Drug-target Binding Affinity Prediction}\label{DTI prediction}

In this study, we look on drug-target prediction as a regression task by predicting the binding affinity values.
With the representation learned from the previous sections, we can integrate all the information  from drugs and targets to predict the binding affinity. In brief, we concatenate all the representations and feed them to three fully-connected dense layers to output the affinity value.
More precisely, for the GAT block, we use two graph attention layers to update the node vectors in a graph considering their neighbor nodes. For the CNN block, we use three consecutive 2D-convolutional layers. And for  the BiGRU block, we use  one BiGRU layer. Besides, we use Rectified Linear Unit (ReLU) \cite{nair2010rectified} as the activation function, which has been commonly adopted in deep learning research. Given a set of drug-target pairs and the ground-truth affinity values in the training dataset, we can use the mean square error (MSE) as the loss function: $\mathcal{L}_{MSE} = \frac{1}{N} \sum_{i=1}^N {(\hat{y}_i - y_i)^2}$, where $\hat{y}_i$ is the predicted value, $y_i$ is the ground-truth value, and $N$ represents the number of drug-target pairs.

\section{Experiments}\label{sec:experiment}


In this part, we first describe the experimental settings (Section \ref{sec:setting}). Then, we compare our proposed method with  state-of-the art models (Section \ref{sec:compare}). Besides, we conduct more experiments to analyze our model including the prediction performance and sensitiveness (Section \ref{sec:analysis}). Finally, we conduct an ablation study to investigate the effectiveness of main strategies suggested in the paper (Section \ref{sec:ablation}).

\subsection{Experimental Setup} \label{sec:setting}
\subsubsection{Datasets}

Following prior works \cite{ozturk2018deepdta,He2017SimBoost}, we employed widely-used datasets that are specialized for DTA prediction:
\begin{itemize*}
\item The \textbf{Davis} dataset, which contains 68 drugs and 442 targets with 30,056 drug-target interactions.
\item The \textbf{KIBA} dataset, which originally comes from a method named Kinase Inhibitor BioActivity (KIBA), and it introduces KIBA scores with integration of the statistic information of $K_d$, $K_i$ and $IC_{50}$ into a single bioactivity score for drug-target interaction. The dataset contains 2,111 drugs and 229 targets with 118,254 interactions after processing~\cite{ozturk2018deepdta}.
\end{itemize*}
\textcolor{black}{We randomly split the datasets into 6 subsets with the equal size, and used five of them for training and the remaining one for testing. For Davis dataset, we use the $K_d$ values transformed into log space, $pK_d$, as the binding affinity value. For KIBA dataset, it integrated from multiple sources (i.e., $K_i$, $K_d$, and $IC_{50}$) into a bioactivity matrix, we use the value (i.e., KIBA-values) in matrix as the binding affinity value.}





\subsubsection{Evaluation Metrics}
We used four metrics commonly used in regression task (recall Section \ref{DTI prediction}) to evaluate the performance. They include:  Mean Squared Error (MSE), Concordance Index (CI), $r_{m}^{2}$, and Area Under Precision Recall (AUPR) score.

\textbf{MSE} has been defined in the previous section as the objective of {DeepGS}. \textbf{CI} \cite{ozturk2018deepdta} measures whether the predicted binding affinity values rank the corresponding drug-target interactions in the same order as the \textcolor{black}{ground-truth} does. \textcolor{black}{It is computed as
$CI= \frac{1}{Z} \sum_{y_{i}>y_{j}} \zeta(f_i - f_j)$ and $\zeta(b)=\left\{1, b>0; 0.5, b=0; 0, b<0 \right\}$, where} $Z$ is a normalization constant that equals the number of drug-target pairs with different binding affinity values. More specifically, when $y_i>y_j$, a positive score is given if and only if the predicted $f_i>f_j$. Here, $\zeta(b)$ is a step function.

The metric \bm{$r_{m}^{2}$} is used to evaluate the external prediction performance of QSAR (Quantitative Structure-Activity Relationship) models. A model is acceptable if and \textcolor{black}{only if $r_{m}^{2} \ge$ 0.5. And $r_{m}^{2}= r^{2} * (1 - \sqrt{r^{2}-r_{0}^{2}})$, where} $r^{2}$ and $r_{0}^{2}$ represent the squared correlation coefficient values between the observed and predicted values with and without intercept, respectively.

The \textbf{AUPR} score is widely used for binary classification. A commonly used binding affinity value is defined based on the logarithm of $K_d$ as \textcolor{black}{$pK_d= -log10(\frac{K_d}{1e9}) \nonumber$, where} $K_d$ refers to the dissociation value~\cite{Tang2014Making}. Here, we transformed the datasets into binary datasets with predefined thresholds. We followed the prior work \cite{ozturk2018deepdta} to select $pK_d$ value of 7 and 12.1 as threshold for the Davis and KIBA dataset, respectively.

\subsubsection{Baseline Methods}
We compared DeepGS against the following state-of-the-art models:

\begin{itemize*}
\item \textbf{KronRLS \cite{Pahikkala2015Toward}:} This approach is based on Kronecker Regularized Least Square \textcolor{black}{ (http://staff.cs.utu.fi/~aatapa/software/RLScore/)}. It aims to minimize the objective function, $J(f)= \sum_{i=1}^m (y_i-f({x_i}))^2 + \lambda {\parallel{f}\parallel}_k^2$,
    where $x_i$ ($i$=1,...,$m$) is a set of training input features, $f$ is a non-linear function, $y_i$ represents their corresponding real-valued labels, and $\lambda>0$ is a pre-defined regularization parameter, ${\parallel{f}\parallel}_k^2$ is the norm of $f$ with kernel $k$. 
\item \textbf{SimBoost \cite{He2017SimBoost}:} 
    This baseline constructs three kinds of features and trains a gradient boosting machine \cite{Higgs2015Chen} model to represent the nonlinear associations between the input features and the binding affinities. 
\item \textbf{DeepCPI \cite{tsubaki2018compound}
    :} This baseline is originally designed for CPI/DTI prediction, and cannot be used directly for DTA task. 
    Here, we need to change it to a regression task. Specifically, we replaced the loss function of \textit{cross\_entropy} with \textit{MSE}, and set the dimension of output layer to 1. The rest is consistent with the original paper. 
\item \textbf{DeepDTA \cite{ozturk2018deepdta}
    :} 
    {DeepDTA}  trains two 3-layer CNNs with label/one-hot encodings of compound and protein sequences to predict DTA task. Their CNN model consists of two separate CNN blocks to learn the features from SMILES strings of compounds and protein sequences, respectively. The representations of drugs and targets are concatenated and passed to a fully connected layer for DTA prediction.
\end{itemize*}

\textcolor{black}{As for KronRLS and SimBoost, they both use PubChem clustering server for drug similarity and Smith-Waterman for protein similarity computation; For DeepDTA, the input for Davis dataset consists of (85, 128) and (1200, 128) dimensional matrices for the compounds and proteins, respectively, and with a (100, 128) dimensional matrix for the compounds and a (1000, 128) dimensional matrix for the proteins for KIBA dataset. The other settings are kept as the same as the original paper.}

\subsubsection{Implementation Details}
For {Smi2Vec}, we used an embedding layer with 100 dimensions to represent the symbols in SMILES sequences, and for {Prot2Vect} we used 100-dimensional pre-trained representations for the biological words. As a result, we constructed matrices with (100, 100) and (2000, 100) dimensions for drug and target, respectively. In our experiments, when the molecular graph was   used, we  employed the RDKit \cite{landrum2006rdkit} software to convert the textual representation in SMILES format to a   graph representation.
For the GAT block, \textcolor{black}{we set the number of heads to 10, and} it was implemented using pytorch\_geometric \textcolor{black}{ (https://github.com/rusty1s/pytorch\_geometric)}, and we set the same radius $r=2$ as in  \cite{tsubaki2018compound}. For the BiGRU block, we set the size of input and hidden layer to 100. For the CNN block, we set the size of kernel to 23. \textcolor{black}{Note that, we performed grid search over a combination of the hyper-parameters to determine the settings. The detailed settings are summarized in  Table \ref{settings}.} And we obtained a high performance of the proposed framework with a relatively small range on hyper-parameter tuning. The proposed framework was implemented using PyTorch with Tensorflow \cite{Abadi2016Tensorflow} backend and ADAM optimization. Our experiments were run on Linux 16.04.10 with Intel(R) Xeon(R) CPU E5-2678 v3@2.50GHz and GeForce GTX 1080Ti (11GB).
\begin{table}
\caption{\textcolor{black}{The detailed training settings of DeepGS. The length of SMILES/protein sequence has three various settings, which are used to study the impact of SMILES/protein sequence length. }}
\scriptsize
\begin{center}
\begin{tabular}{c c c c}
\toprule
\textbf{Parameter} & \textbf{Setting} & \textbf{Parameter} & \textbf{Setting}\\
\midrule
Radius r                   &  2            & Layer of CNN           &   3        \\
N-gram                     &  3            & Layer of BiGRU         &   1          \\
CNN kernel size            & 23                 & Learning rate (lr)     &   1e-4   \\
Length of SMILES sequence  & 50, {100}, 500       & lr decay               &  0.9  \\
Length of protein sequence & 500, {1000}, 2000    & Decay interval         & 20  \\
Vector dimension           & 32         & Weight decay           & 1e-5   \\
Window size                & 11         & Epoch                  &   100   \\
Depth in GAT               &   2        & Batchsize              &   1    \\

\bottomrule

\end{tabular}

\label{settings}
\end{center}
\end{table}
\vspace{-2ex}
\begin{figure*}[hb]
\centering
\includegraphics[width=1.65in]{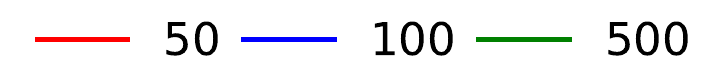}\\
\includegraphics[width=1.65in]{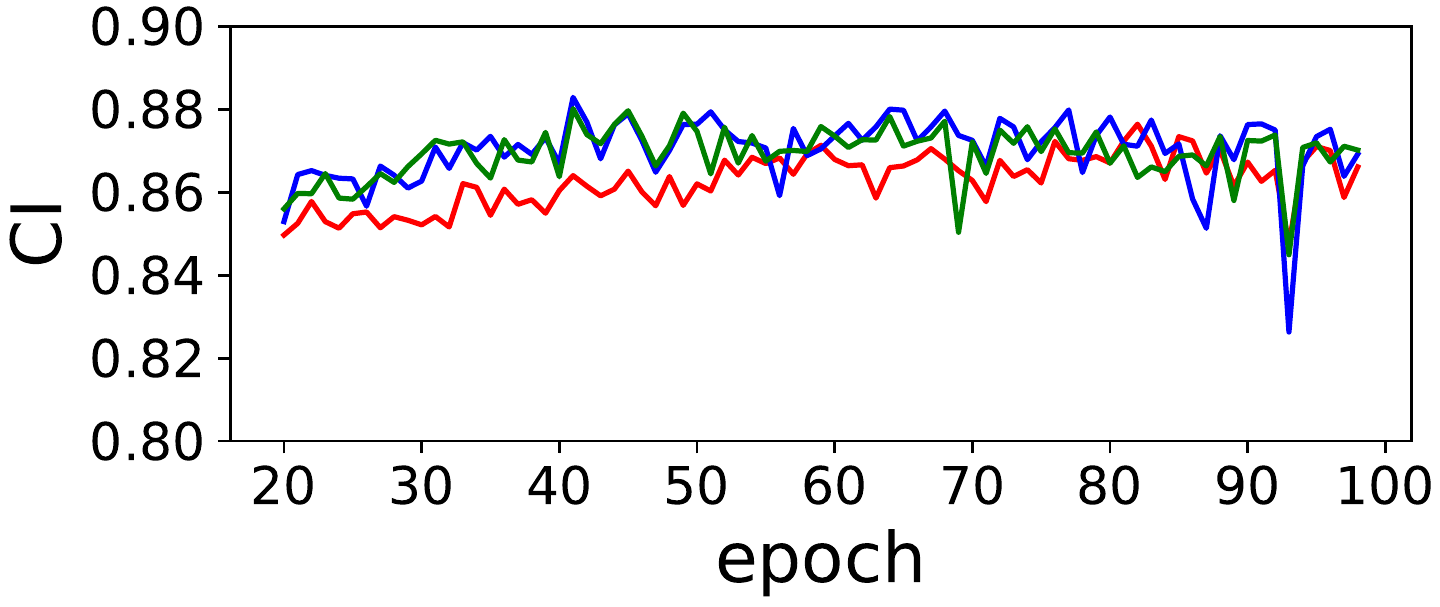}
\includegraphics[width=1.65in]{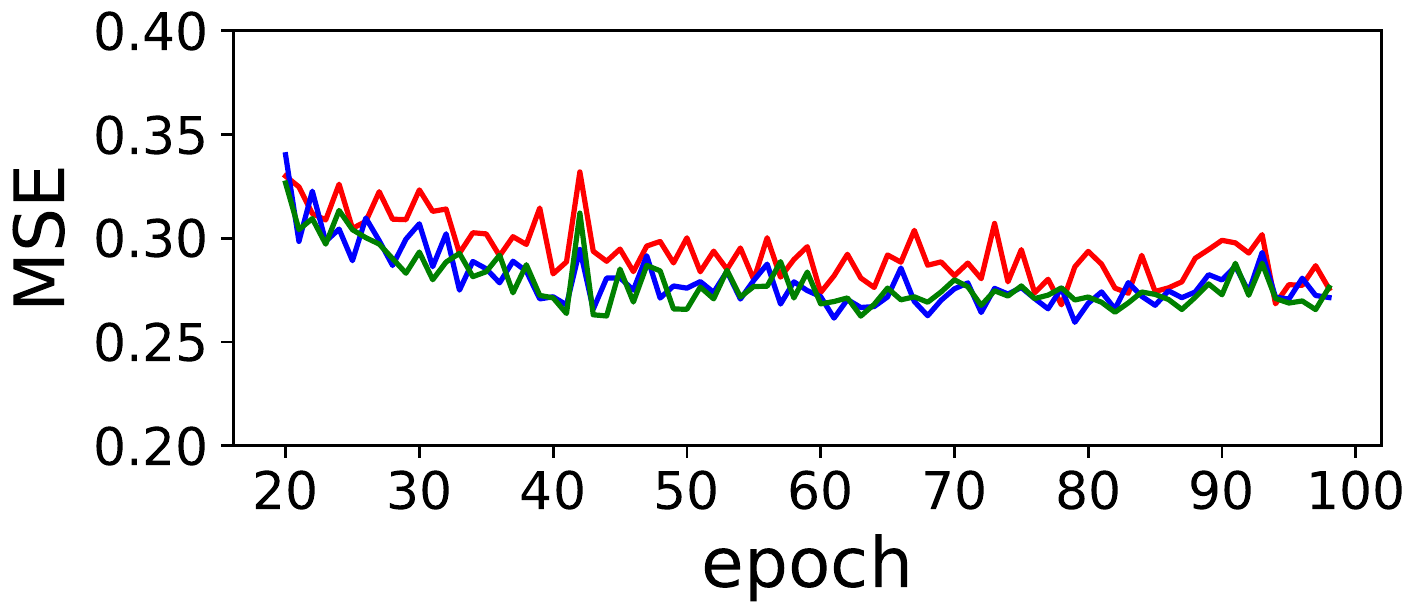}
\includegraphics[width=1.65in]{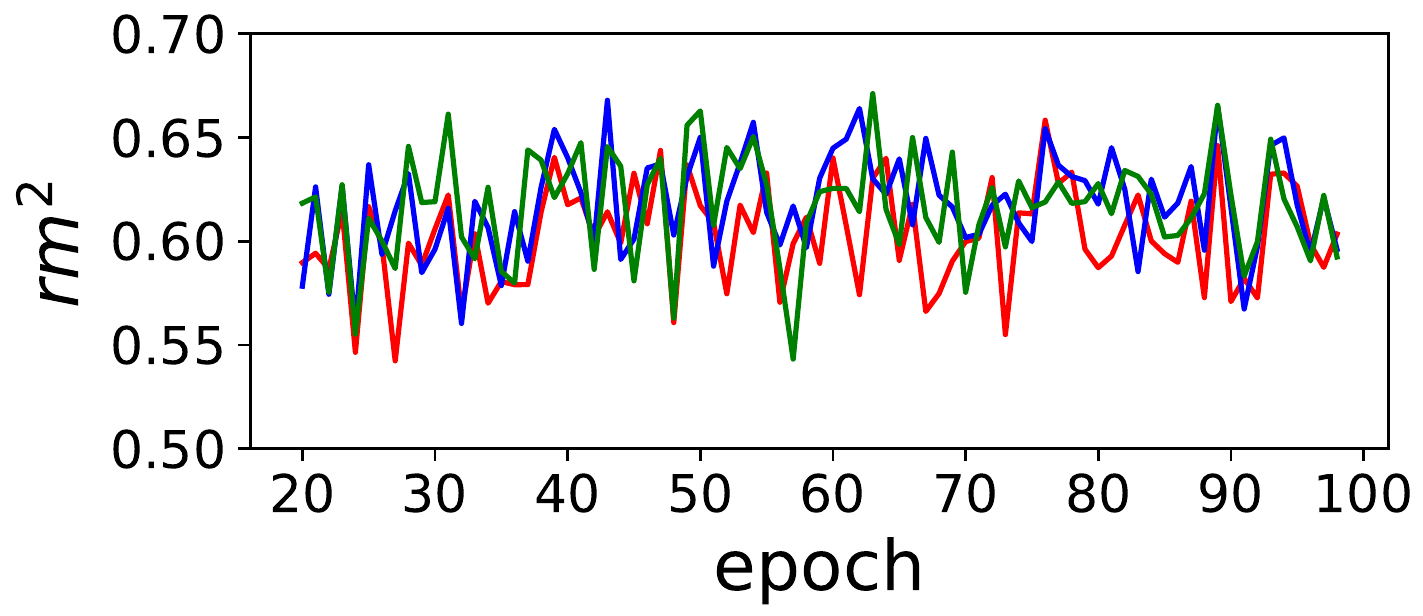}
\includegraphics[width=1.65in]{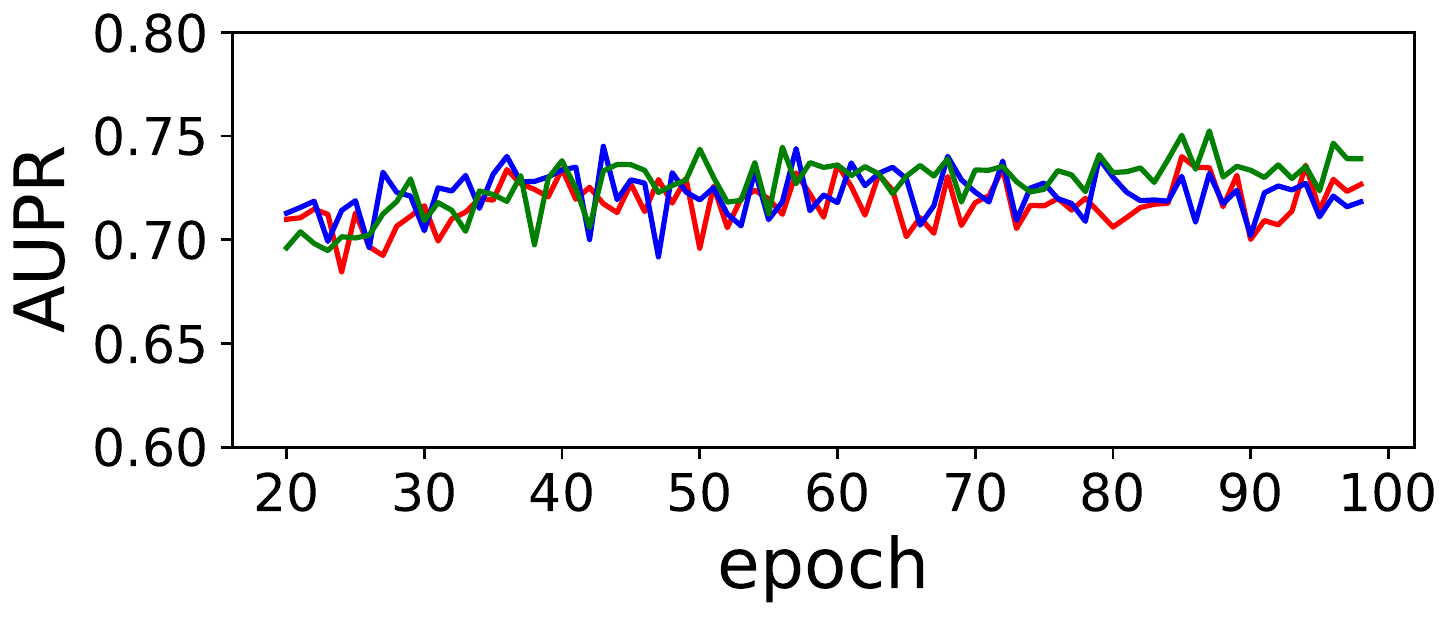}
\vspace {-2ex}
\caption{CI, MSE, $r_m^2$, and AUPR vs. $L_{ds}$ (drug SMILES sequences in the Davis dataset).}
\label{effect1}
\end{figure*}

\subsection{Comparison Results}\label{sec:compare}
To examine the competitiveness of the proposed model, we compared {DeepGS} with state-of-the-art models (including classic and   deep learning models) used for DTA prediction. Table \ref{CI-Davis} reports the average CI, MSE, $r_{m}^{2}$ and AUPR scores on the Davis   dataset.

\begin{table}
\scriptsize
\caption{The average CI, MSE, $r_{m}^{2}$ and AUPR scores  on the Davis dataset. The results of {KronRLS}, {SimBoost} and {DeepDTA} are reported from \cite{ozturk2018deepdta}.}
\begin{center}
\scalebox{0.9}{
\begin{tabular}{c c c c c c c}
\toprule
Method            & Drugs         & Targets          & CI             & MSE            & $r_{m}^{2}$    & AUPR\\
\midrule
KronRLS           & Pubchem       & S-W              & 0.871          & 0.379          & 0.407          & 0.661\\
SimBoost          & Pubchem       & S-W              & 0.872          & 0.282          & 0.644          & 0.709\\
DeepCPI           & GNN           & CNN (Embedding)  & 0.867          & 0.293          & 0.607          & 0.705\\
DeepDTA           & CNN           & CNN (One-hot)    & 0.878          & 0.261          & 0.630          & 0.714\\
\textbf{DeepGS}   & GAT+Smi2Vec   & CNN (Prot2Vec)   & \textbf{0.882} & \textbf{0.252}    & \textbf{0.686} & \textbf{0.763}\\
\bottomrule
\end{tabular}}

\label{CI-Davis}
\end{center}
\vspace{-2ex}
\end{table}
\vspace{-2ex}

From this table, we can see that, on the whole classic methods such as {KronRLS} perform worse than deep learning-based methods. This is because classic methods rely heavily on hand-crafted features and the similarity matrices of drugs and targets. In contrast, deep learning-based approaches capture more information via automatic feature engineering with CNN and GNN. In addition, we find that our method performs better than other two deep learning-based methods. The reason could be that (\romannumeral 1) compared to {DeepCPI}, our method jointly considers topological structures and  local chemical context, which is benefit to the performance; (\romannumeral 2) compared to {DeepDTA}, we incorporate  GAT model to obtain the topological information of drug and advanced embedding techniques which bring more  contextual information than one-hot vectors for modeling both drugs and targets.

Overall, this set of experiments demonstrate that our proposed method {DeepGS} outperforms all these baselines in all metrics. This is a very encouraging result. It is worth noting that, although the improvements seem to be small at the first glance, it is essentially a non-trivial achievement in terms of DTA prediction.



\begin{table}
\scriptsize
\caption{\textcolor{black}{The average CI, MSE, $r_{m}^{2}$ and AUPR scores on the KIBA dataset. The results of {KronRLS}, {SimBoost} and {DeepDTA} are reported from \cite{ozturk2018deepdta}.}}
\begin{center}
\scalebox{0.9}{
\begin{tabular}{c c c c c c c}
\toprule
Method            & Drugs        & Targets            & CI             & MSE            & $r_{m}^{2}$     & AUPR\\
\midrule
KronRLS           & Pubchem      & S-W                & 0.782          & 0.411          & 0.342           & 0.635\\
SimBoost          & Pubchem      & S-W                & 0.836          & 0.222          & 0.629           & 0.760\\
DeepCPI           & GNN          & CNN (Embedding)    & 0.852          & 0.211          & 0.657           & 0.782\\
DeepDTA           & CNN          & CNN (One-hot)      & \textbf{0.863} & 0.194          & 0.673           & 0.788\\
\textbf{DeepGS}   & GAT+Smi2Vec  & CNN (Prot2Vec)     & 0.860          & \textbf{0.193} & \textbf{0.684}  & \textbf{0.801}\\
\bottomrule

\end{tabular}
}
\label{CI-KIBA}
\end{center}
\vspace{-2ex}
\end{table}

Besides the comparison on the Davis dataset, we also conduct the comparison  on the KIBA dataset. Table \ref{CI-KIBA} shows the comparison results. It can be seen that, the overall performance tendency is similar to that on the Davis dataset. For example, the performance of KronRLS is inferior to that of deep learning-based approaches, the performance of {DeepCPI} is inferior to that of {DeepDTA}, and our method exhibits better performance on almost all these metrics. This further demonstrates the competitiveness of DeepGS. Note that, in terms of CI metric, our method still has the comparable performance to DeepDTA, since the  value of our method is only slightly smaller than that of DeepDTA. The possible reason is that, the KIBA dataset comes from multiple sources (\emph{e.g.,} $K_i$, $K_d$ and $IC_{50}$, recall Section \ref{sec:setting}), the data heterogeneity in KIBA dataset may make a negative effect on the CI metric of our  model.

\subsection{Model Analysis}\label{sec:analysis}
In this section, we conduct more experiments to analyze our model. In the first experiment, we further examine the prediction performance of our model based the predicted value $(p)$ and measured value $(m)$. In the second experiment, we examine the sensitiveness of our model by using various sequence lengths.

\begin{figure}[h]
\centering
\includegraphics[width=0.48\linewidth]{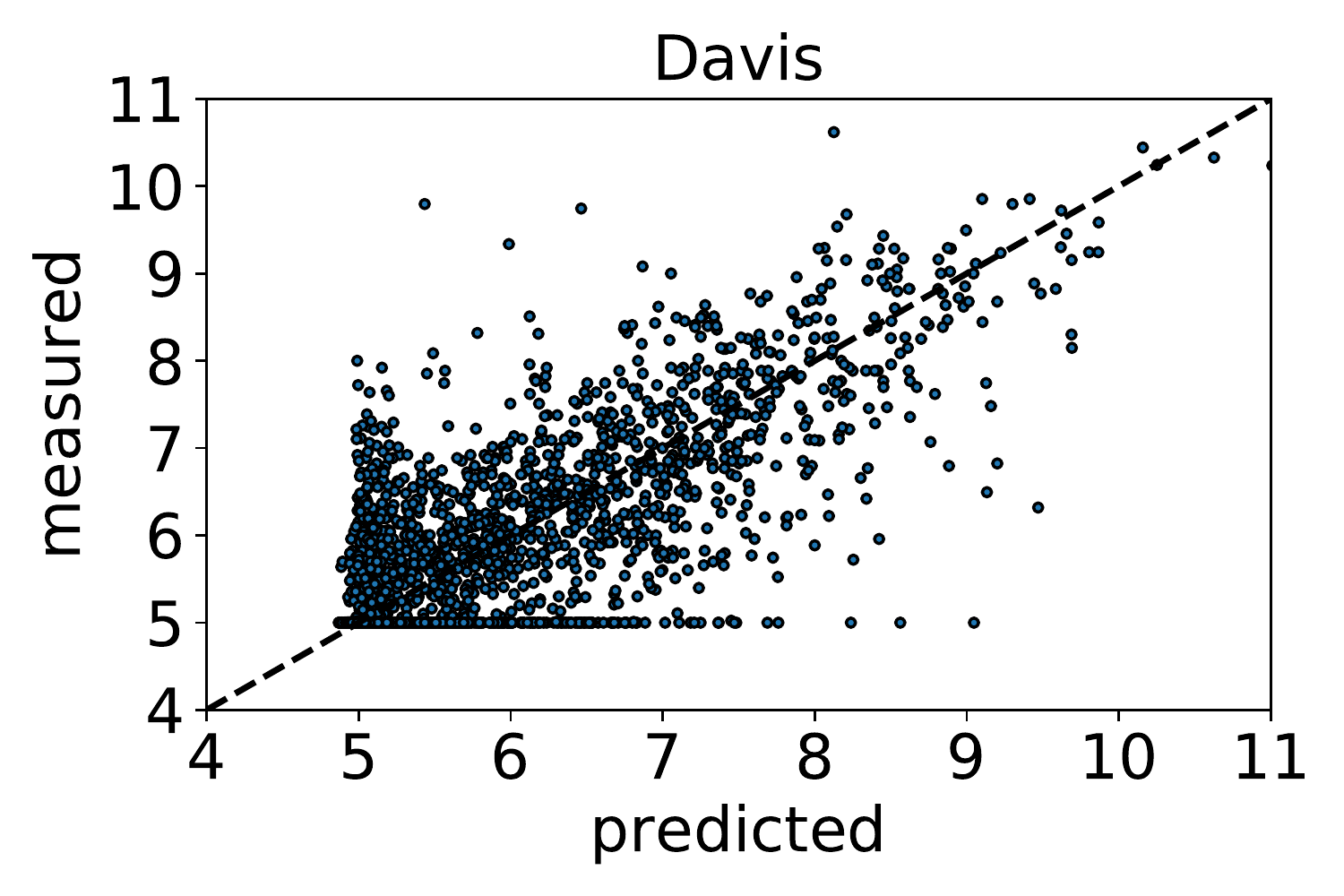}
\includegraphics[width=0.48\linewidth]{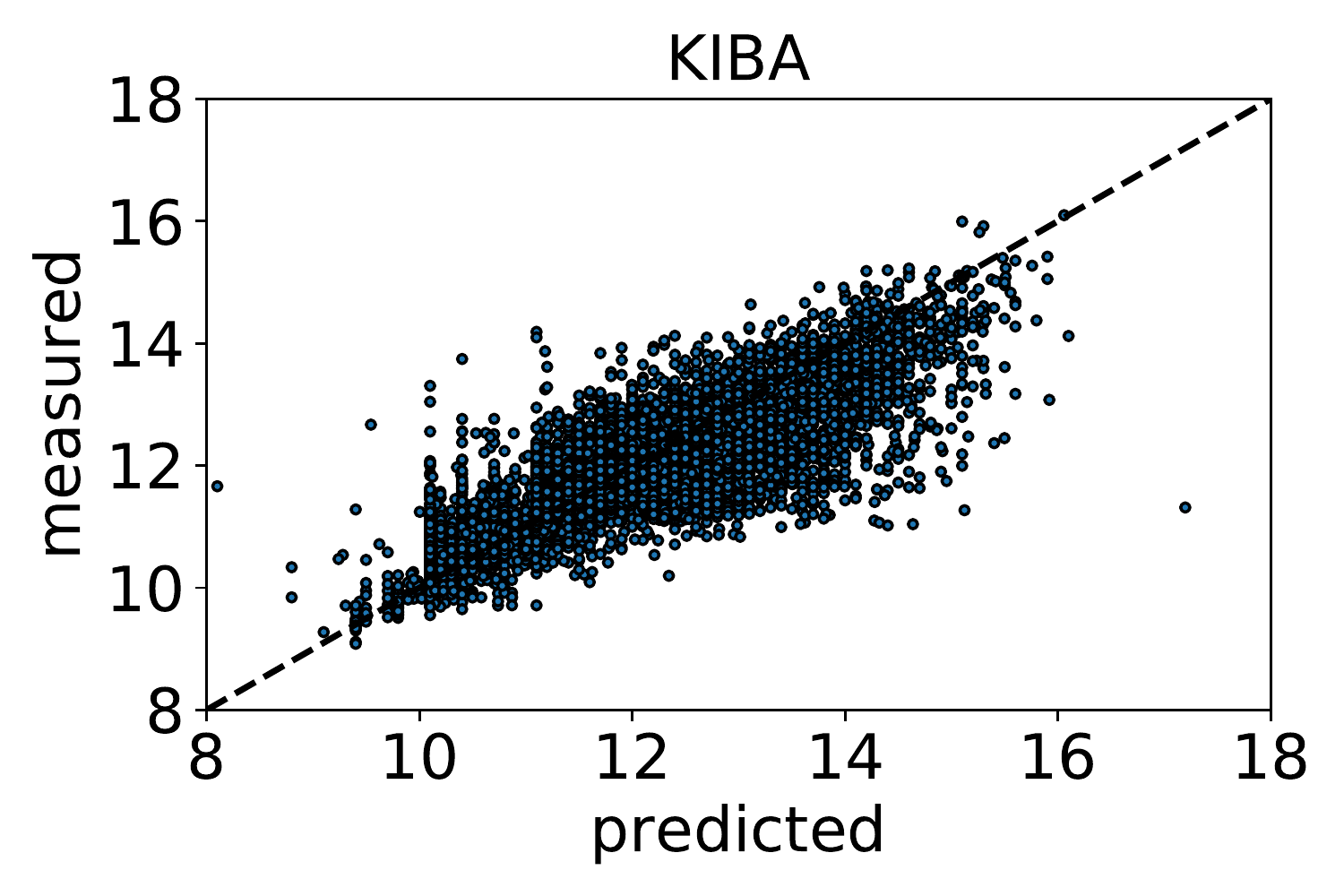}
\vspace{-2ex}
\caption{\textcolor{black}{Predictions from DeepGS model against measured binding affinity values. The left and right figures plot  the results on the Davis and KIBA datasets, respectively.}}
\label{measured}
\end{figure}

\begin{figure*}
\centering
\includegraphics[width=1.65in]{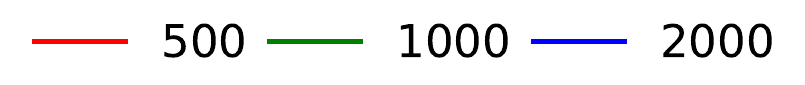}\\
\includegraphics[width=1.65in]{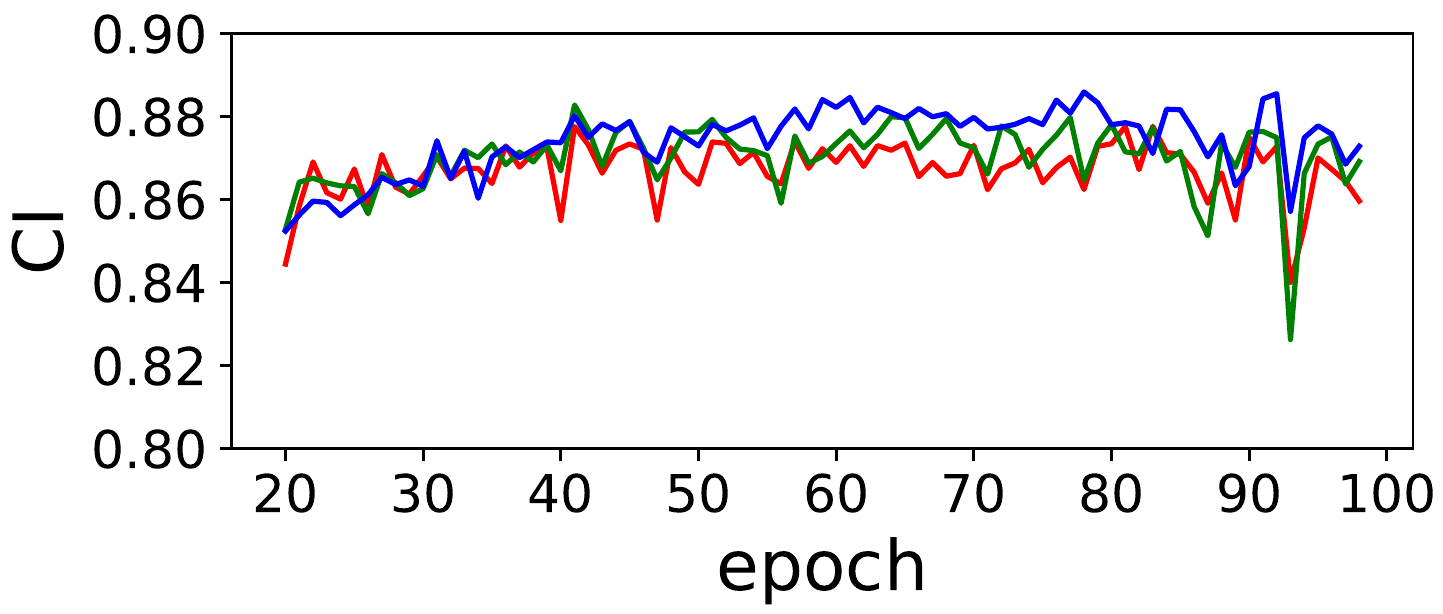}
\includegraphics[width=1.65in]{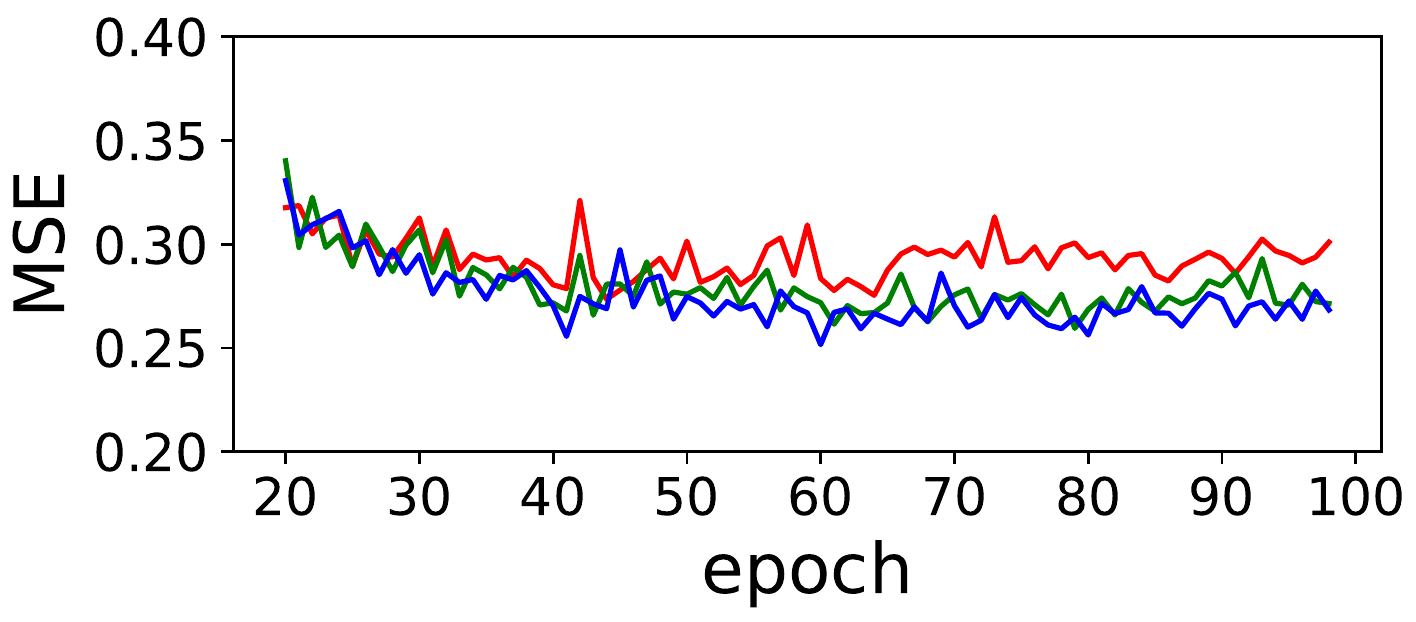}
\includegraphics[width=1.65in]{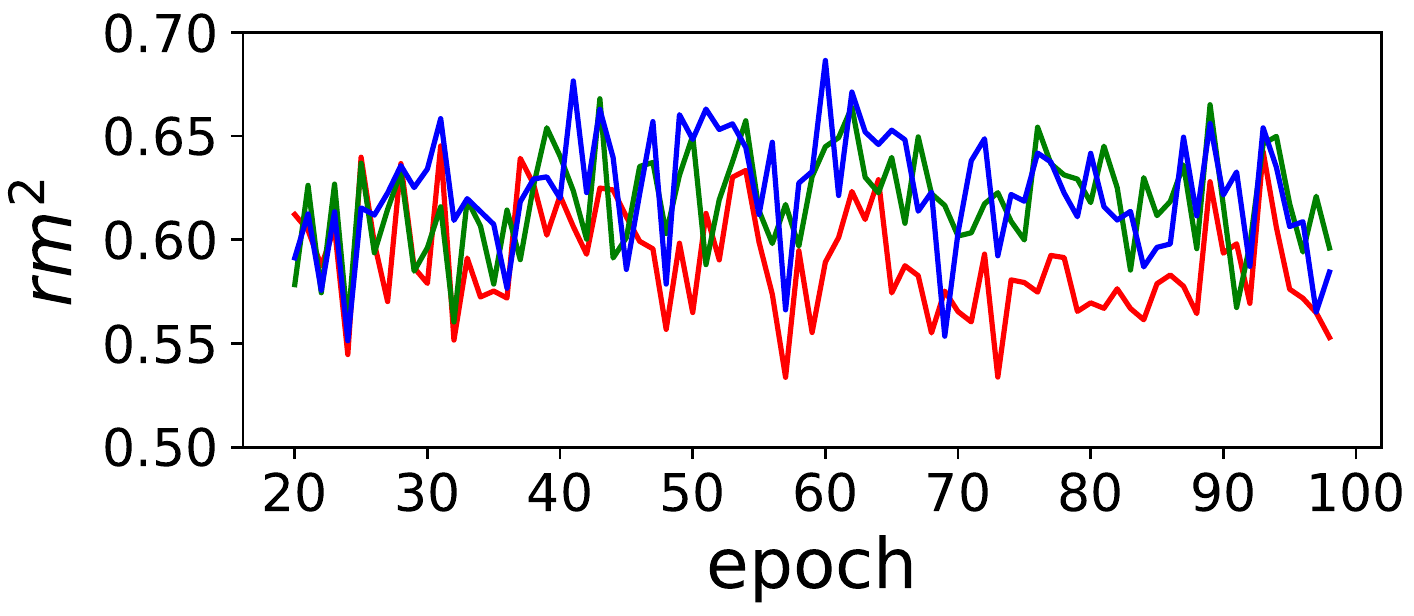}
\includegraphics[width=1.65in]{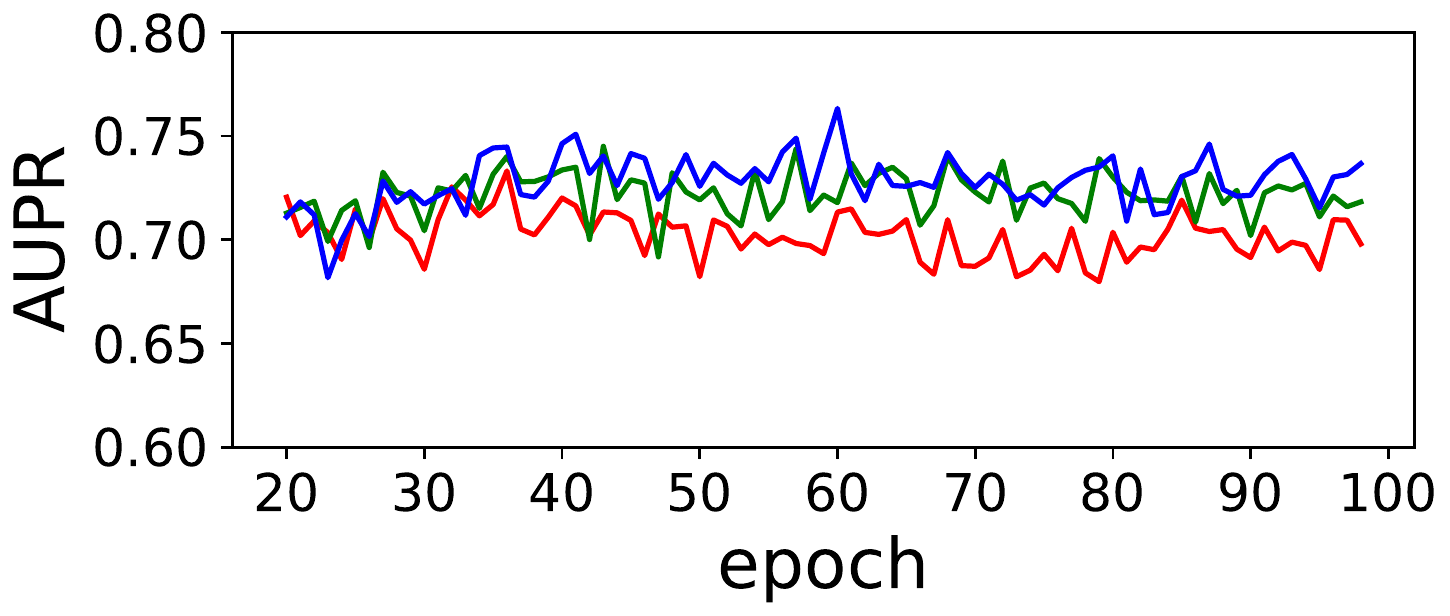}
\vspace {-2ex}
\caption{CI, MSE, $r_m^2$, and AUPR vs. $L_{ps}$ (protein sequences in the Davis dataset).}
\label{effect2}
\end{figure*}

Figure \ref{measured} plots the predicted value $(p)$ and measured value $(m)$ on these two datasets. Note that, a good model should hold that predicted value $(p)$ is close to the measured value $(m)$, and thus the samples should fall close to the dashed ($p=m$) line. One can see that, for the Davis dataset, the dense area of the $pK_d$ value is in the range of 5 to 6 in terms of x-axis. This is because the $pK_d$ value of 5 constitutes more than half of the dataset (i.e., 20,931 out of 30,056, as reported from \cite{ozturk2018deepdta}). In addition, we observe that the dense area of the KIBA score is in the range of 10 to 14 in terms of x-axis. The reason is similar to that for the Davis dataset. Particularly, for both datasets, the samples are close to the dashed ($p=m$) line. This justifies, from another perspective, that the proposed solution has a good prediction performance.

\begin{figure*}[b]
\centering
\includegraphics[width=2.4in]{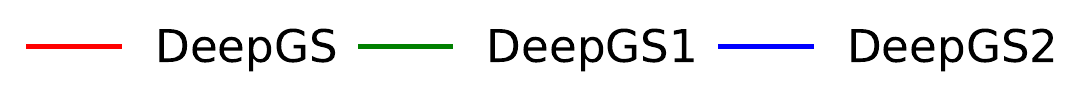} \\
\includegraphics[width=1.65in]{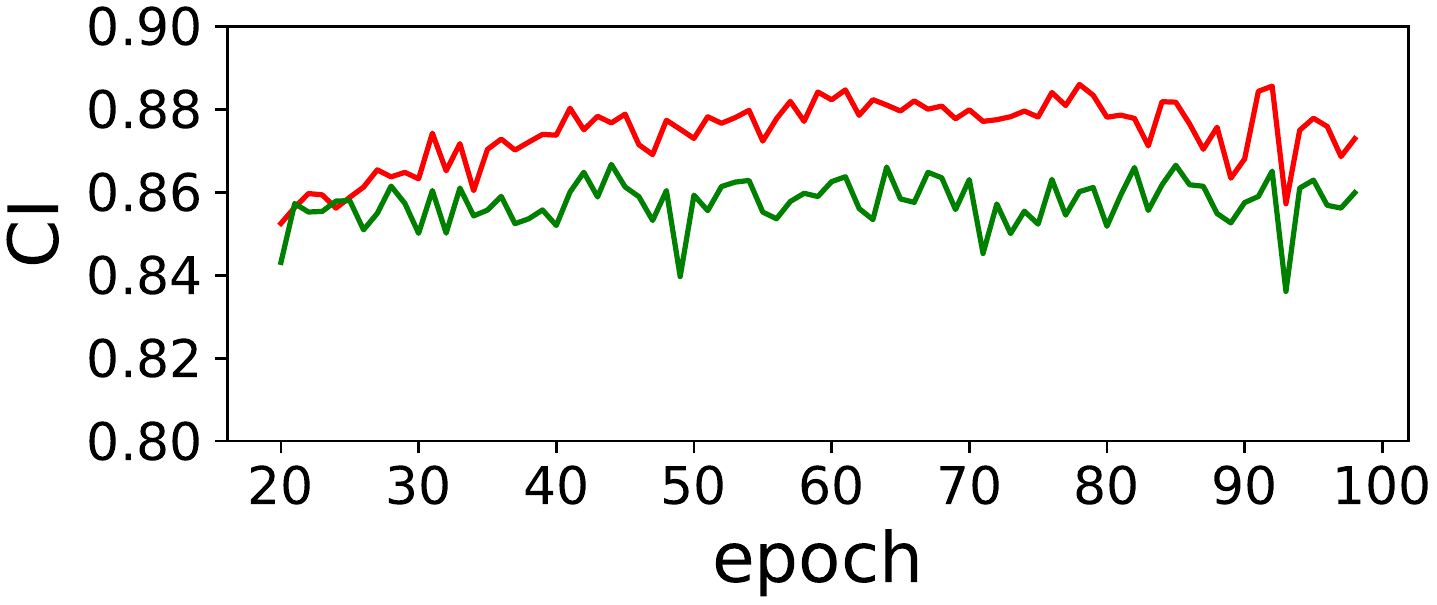}
\includegraphics[width=1.65in]{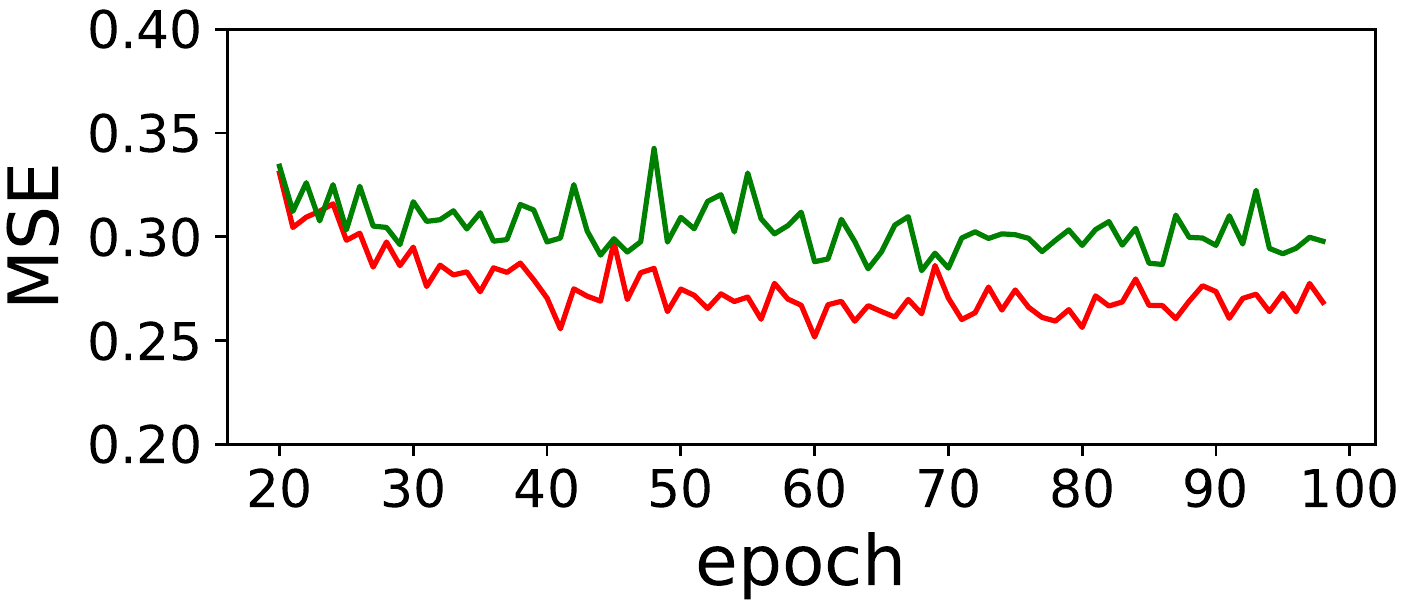}
\includegraphics[width=1.65in]{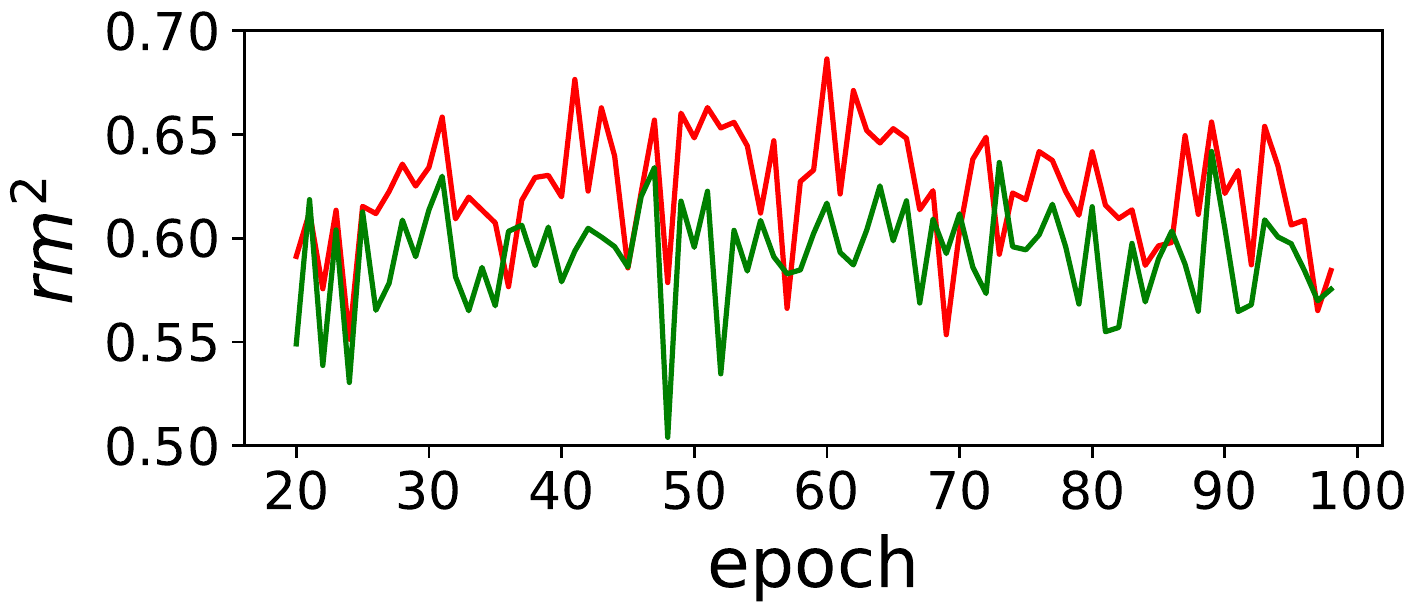}
\includegraphics[width=1.65in]{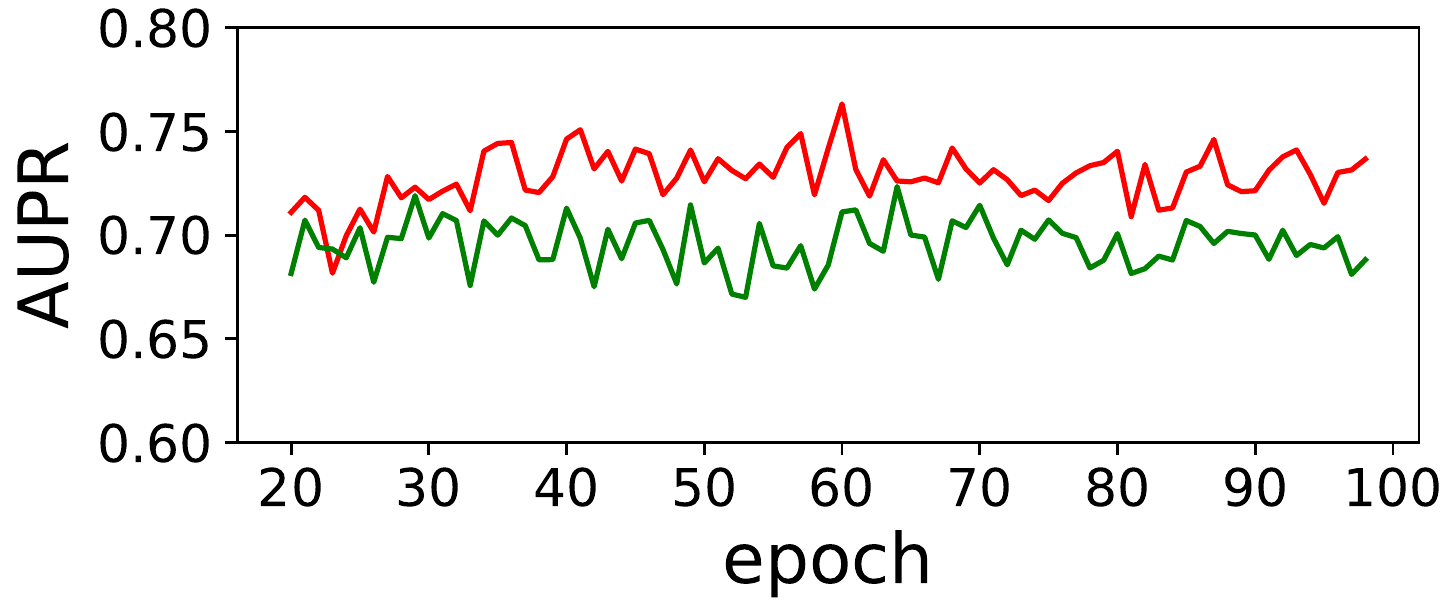}

\includegraphics[width=1.65in]{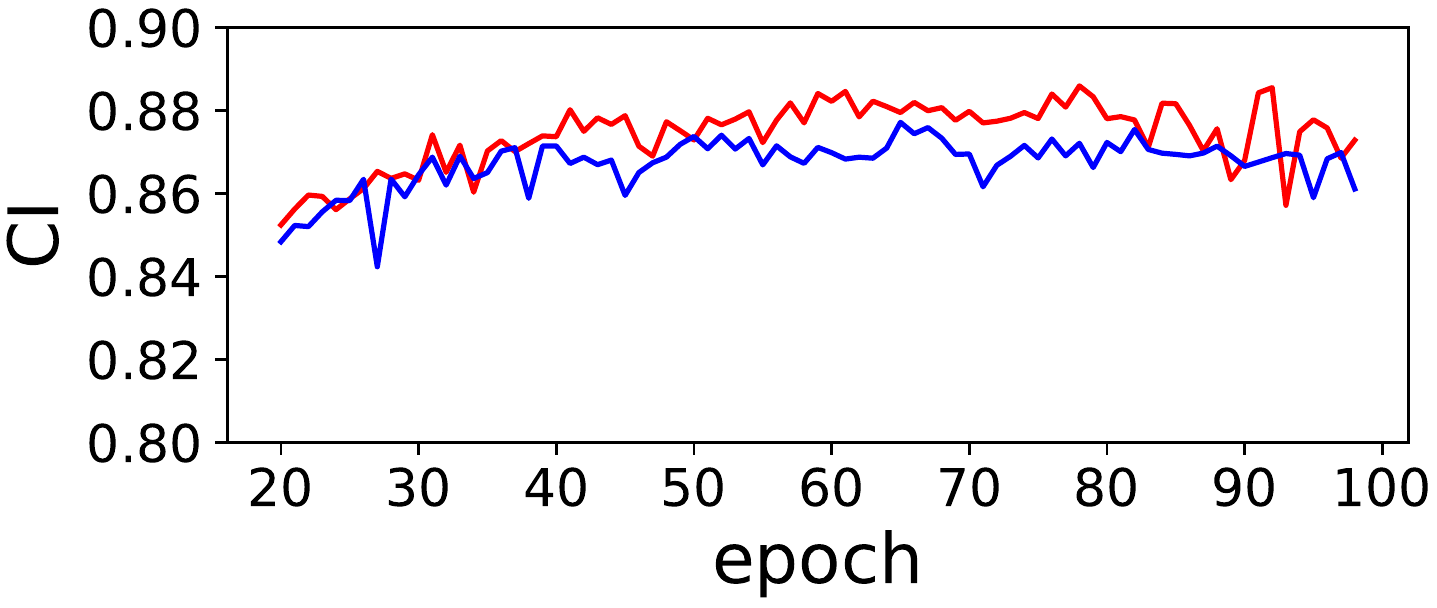}
\includegraphics[width=1.65in]{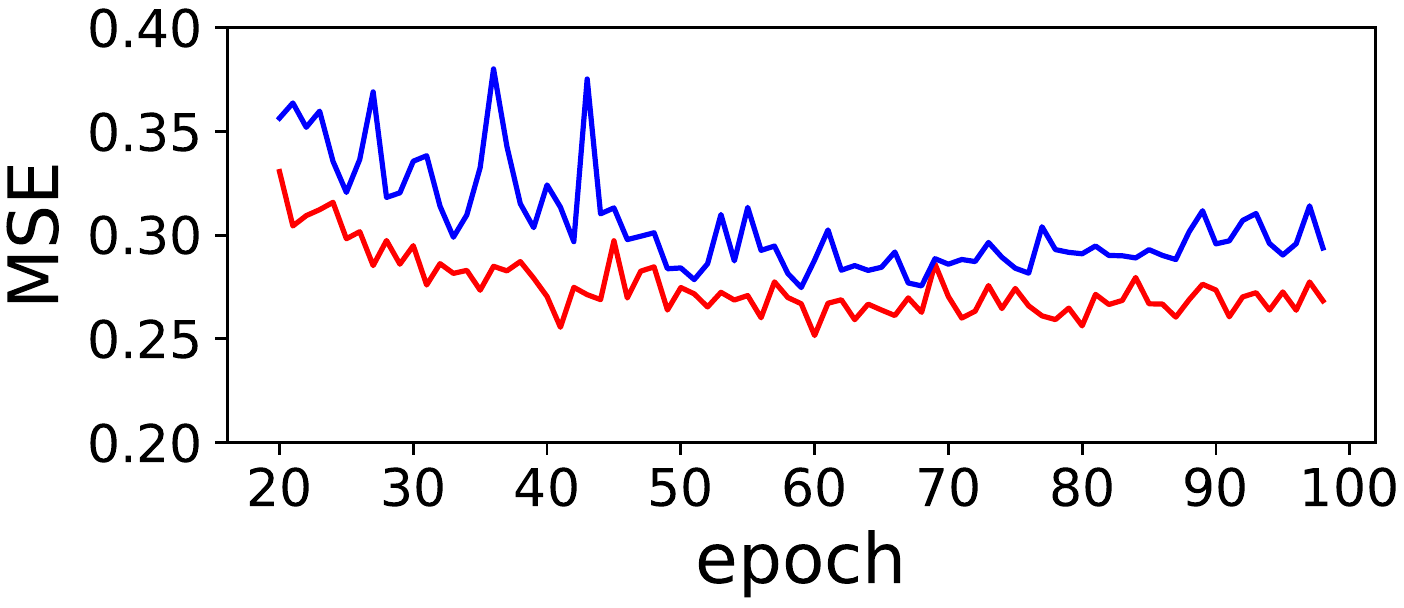}
\includegraphics[width=1.65in]{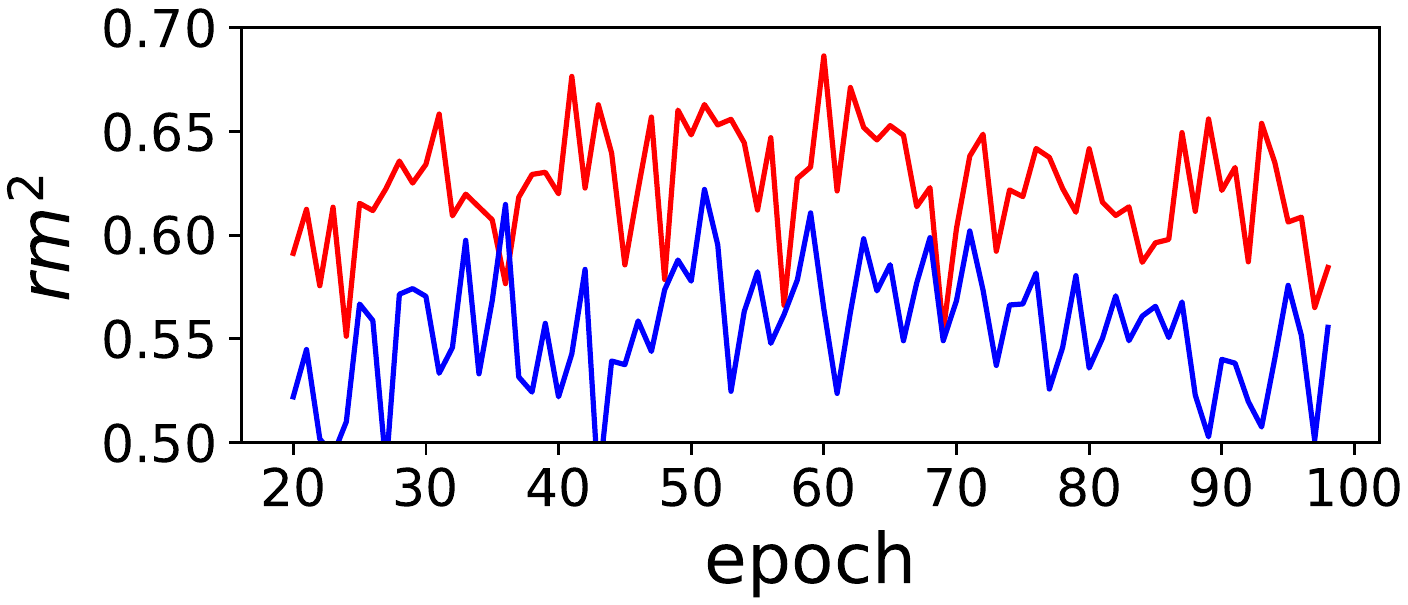}
\includegraphics[width=1.65in]{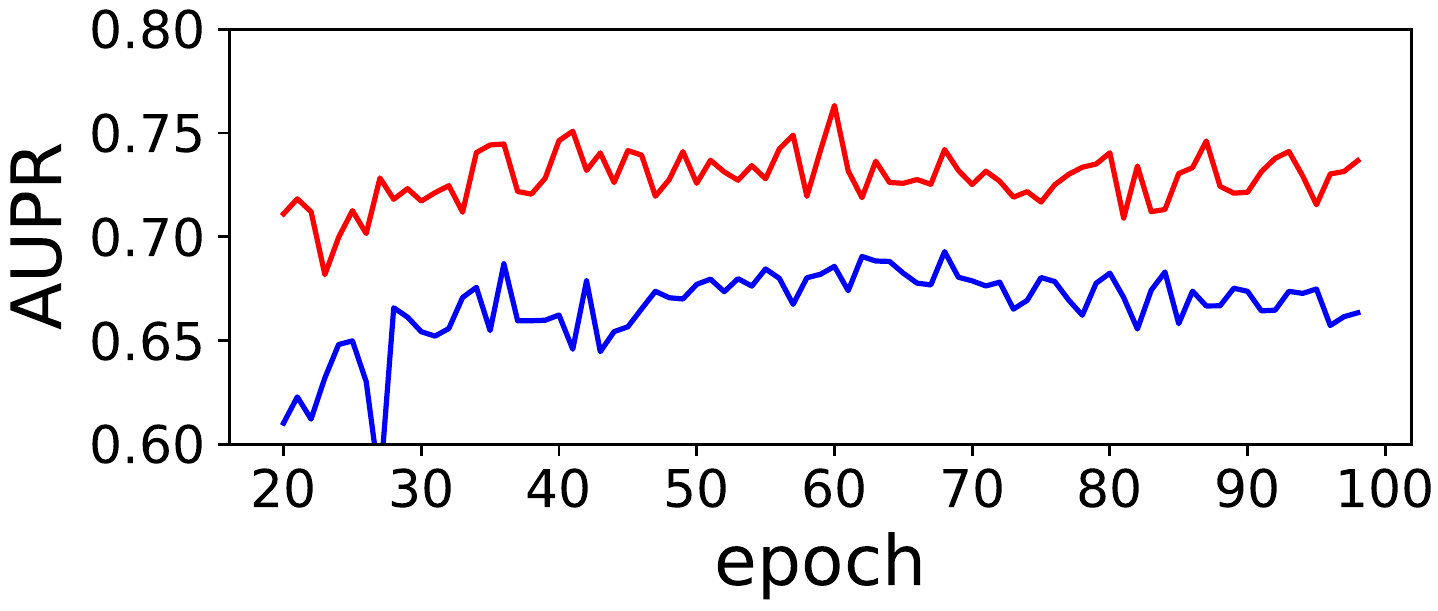}

\vspace {-2ex}

\caption{Ablation study on all metrics for our proposed model and two variants on the Davis dataset. }
\label{DeepGS3}
\end{figure*}

To investigate the sensitiveness of our model, a simple way is to remove some information of the input sequences, and then to test the model's prediction performance. In this paper, we use the following scheme that not only can remove some information of input sequences but also can partially reflect the impact of sequence length. Specifically, we fix the length of the input sequences  at some value, say $l$, when training BiGRU and CNN. To achieve this, we cut the input sequence if the length of the input sequence is longer than $l$, and otherwise we use the zero-padding at the end of the input sequence.
As for the drug's input sequence (i.e., SMILES), we set its length  to [50, \bm{$100$}, 500], as shown in Table \ref{settings}. Note that, the average length of SMILES sequences in the Davis dataset is 64 and the \textbf{bold} refers to the default value.  Correspondingly, we set the length of target/protein's input sequence (i.e., amino acids) to [500, \bm{$1000$}, 2000]. Here, the average length of protein sequences in the Davis dataset is 788. For clarity, we use $L_{ds}$ and $L_{ps}$ to denote the Length of \underline{d}rug \underline{s}equence and that of \underline{p}rotein \underline{s}equence, respectively.

\begin{figure}[h]
\centering
\includegraphics[width=0.48\linewidth]{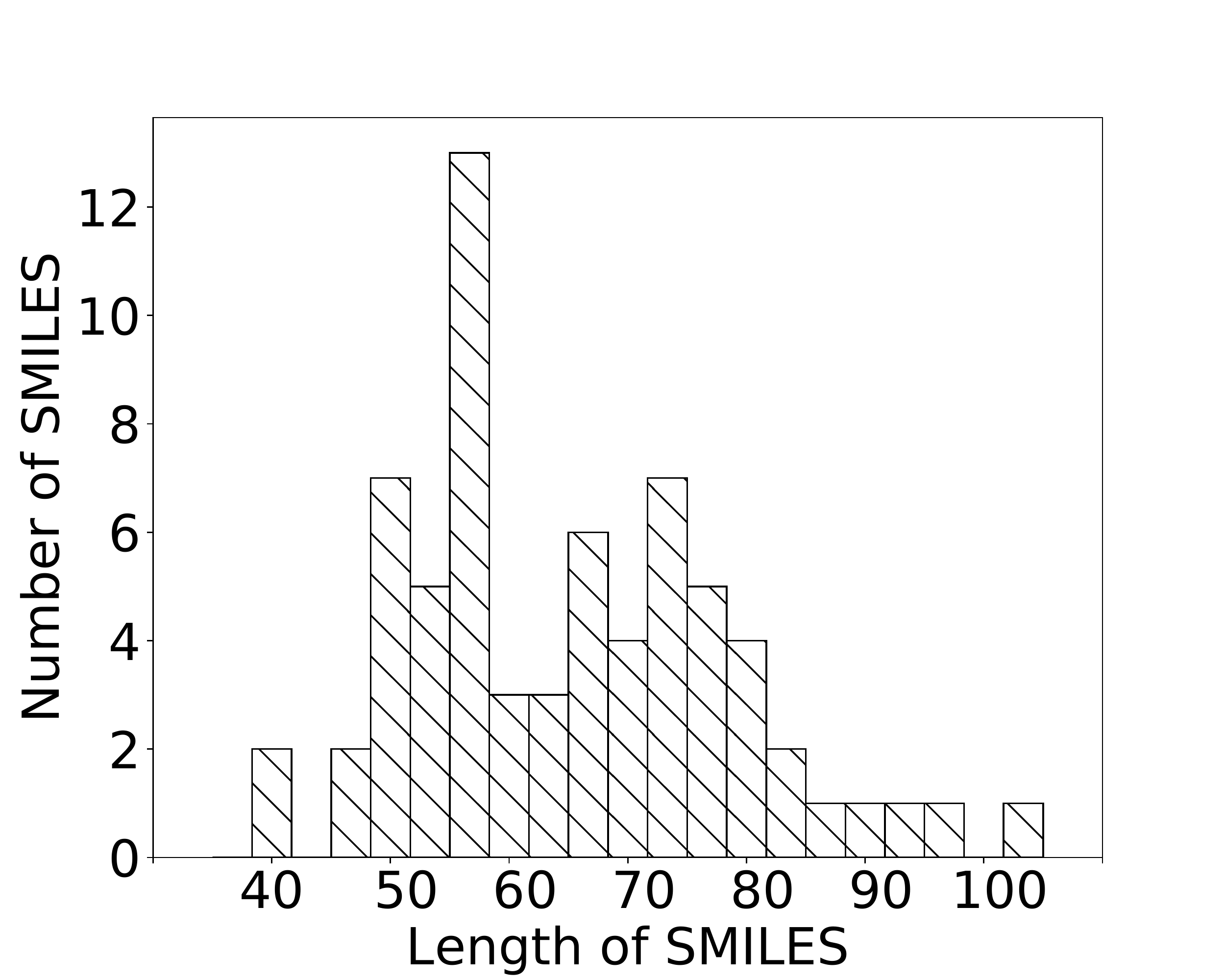}
\includegraphics[width=0.48\linewidth]{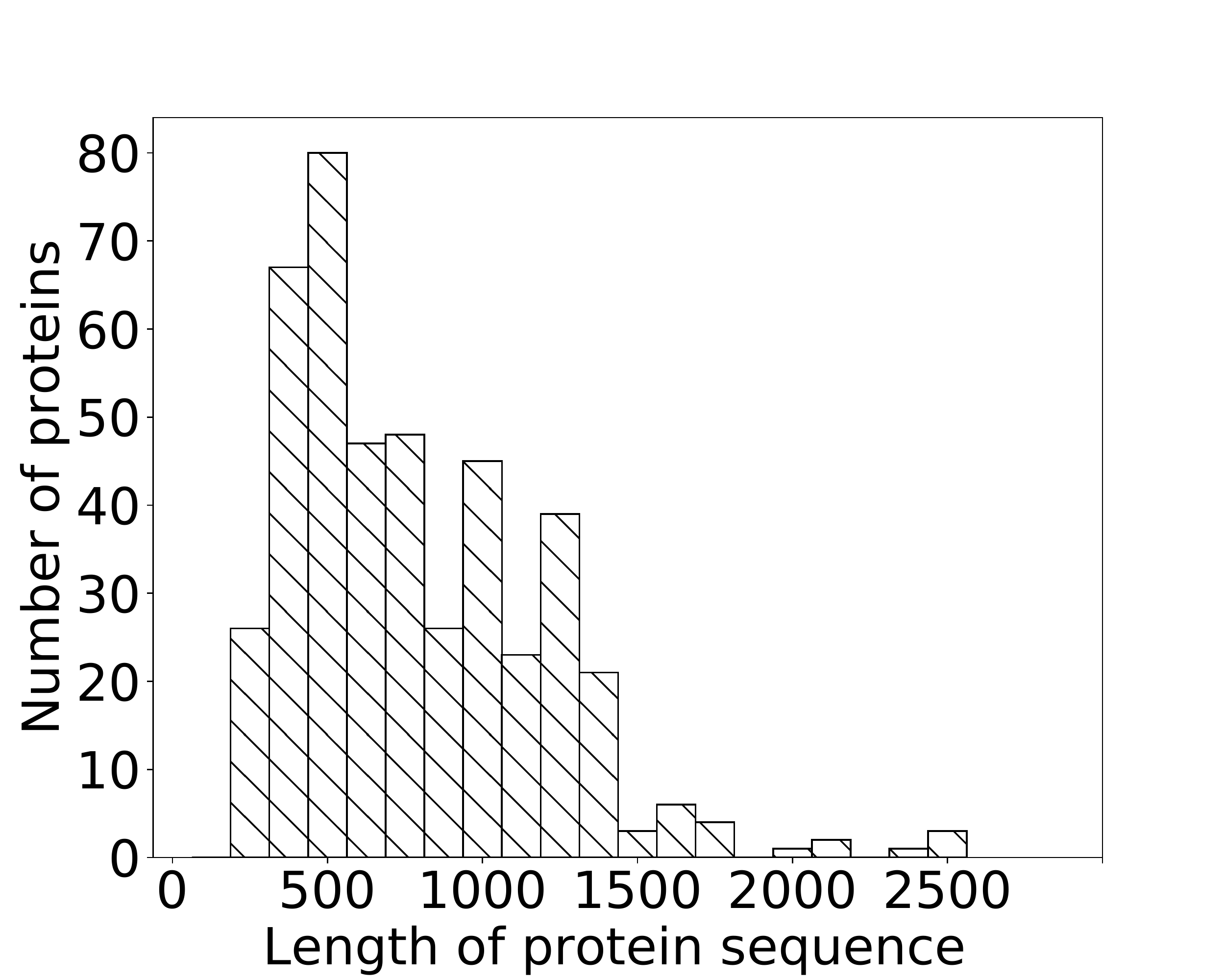}
\vspace{-2ex}
\caption{\textcolor{black}{Sequence length distribution in Davis dataset. The left and right figures refer  to the length distribution of SMILES and protein sequences, respectively.}}
\label{distribution}
\vspace{-3ex}
\end{figure}

Figure \ref{effect1} shows the results when we set $L_{ds}$ to 50, 100 and 500, respectively. Meanwhile, Figure \ref{effect2} shows the results when we set $L_{ps}$ to 500, 1000 and 2000, respectively. We observe that, (\romannumeral 1) the performance gap between   $L_{ds}=100$ and $L_{ds}=500$ is very tiny, and the performance gap between $L_{ps}=1000$ and $L_{ps}=2000$ is not so obvious; (\romannumeral 2) the performance gap between $L_{ds}=50$ and $L_{ds}=100$ can be easily perceived, and  the performance gap between    $L_{ps}=500$ and $L_{ps}=1000$ is obvious. This phenomenon is a little bit strange at the first glance. To dig out the reason behind it, we plot the distribution of sequence lengths, as shown in Figure \ref{distribution}. It can be seen that the lengths of most SMILES sequences are less than 100 and larger than 50. Thus, it is natural that the performance gap between   $L_{ds}=100$ and $L_{ds}=500$ is very tiny, since almost all SMILES sequences do not need to be cut even if $L_{ds}=100$, i.e., few information is missing. However,  when $L_{ds}=50$,  many SMILES sequences may need to be cut, and so the performance degrades. With the similar argument, it is not hard to understand that  the performance gap between $L_{ps}=500$ and $L_{ps}=1000$, since most protein sequences need to be cut when $L_{ps}=500$. The reason for the relatively small  performance gap between $L_{ps}=1000$ and $L_{ps}=2000$ can be inferred with the similar argument. This result may imply that when the sequence length $l$ is set to a value  larger than the average  length of sequences in the dataset, the performance degradation could be trivial.

\subsection{Ablation Study}\label{sec:ablation}
As mentioned before, existing  models for DTA prediction have  leveraged the topological structure to learn the representation for drug and target/protein, while they often ignored the local chemical context. Thereby, this work considers both local chemical context and topological structure to learn the interaction between drugs and targets. More precisely, the core idea of DeepGS is to fully leverage local chemical context, by using advanced embedding techniques, to better learn the drug and target representations. To study the effectiveness of the central idea, we implemented a variant of our model, called DeepGS1. This variant model removes both drug's and protein's local chemical context information obtained by Smi2Vec and Prot2Vec from the framework.  The detailed configuration of DeepGS1 is illustrated in Table \ref{variant}. In the following experiments, we use the same experimental settings mentioned in Table \ref{settings}.

Figure \ref{DeepGS3} shows the comparison results. It can be seen that {DeepGS1} is basically inferior to {DeepGS} in terms of CI, MSE, $rm^2$ and AUPR. These results demonstrate that combining the local chemical context information, which reflects the functionalities of the atoms, is benefit to learning a good representation for drugs and proteins, improving the prediction performance.

In addition to examining  the effectiveness of the central idea, we also conduct another experiment, which   is used to answer the following interesting question. Recall Section \ref{subsec:structureModel}, we develop a GAT-based molecular structure modeling approach. One could argue that, GNN (Graph Neural Network) can also map a graph to a vector that encodes the topological structure of the graph, since recent work \cite{tsubaki2018compound} have showed that GNN can effectively model drugs. To address it, we implemented  another variant of our model, called DeepGS2. Compared to our model, the major difference is that it uses a GNN-based molecular structure modeling approach (cf., Table \ref{variant}).

\begin{table}
\caption{The detailed description of the variant of our model. DeepGS1 is mainly for investigating the local chemical context information, while DeepGS2 is used to justify the choice of GAT used in  our molecular structure modeling  component.}
\scriptsize
\begin{center}
\begin{tabular}{c c c}
\toprule
\textbf{Model} & \textbf{Drug Representation}  & \textbf{Target Representation} \\
\midrule
{DeepGS} & FP+GAT \& Smi2Vec+BiGRU & Prot2Vec+CNN\\
DeepGS1  & FP+GAT  \& one-hot/Label+BiGRU & one-hot/Label+CNN \\
DeepGS2  & FP+GNN \& Smi2Vec+BiGRU & Prot2Vec+CNN \\
\bottomrule

\end{tabular}
\label{variant}
\end{center}
\end{table}

The experimental results are also shown in Figure  \ref{DeepGS3}. We can see that, the variant DeepGS2 is obviously poorer than our model. These results justify our choice in Section \ref{subsec:structureModel}.
The reasons could be two-fold: (\romannumeral 1) the molecular structure may contribute a lot to the prediction performance; and (\romannumeral 2) The changes to the molecular structure modeling approach are sensitive to the model, especially when  local chemical context is also considered in the model.

\section{Related Work}\label{sec:relatedwork}

Drug-target binding prediction has been an interesting topic in drug discovery. Most of previous works focused  on simulation-based methods (i.e., molecule docking and descriptors) or machine learning-based models. For example,
Li et al. \cite{li2015low} proposed a docking method based on random forrest (RF). The RF model was also adopted in \textit{KronRLS} \cite{Pahikkala2015Toward} with a similarity score through the Kronecker product of similarity matrix to improve the predictive performance. To remedy the limitation of linear dependencies in KronRLS, a gradient boosting method was proposed in \textit{SimBoost} \cite{He2017SimBoost} to construct the similarities among drugs and targets.
Although classic methods show reasonable performance in drug-target prediction, they are often computational expensive, or require external expert knowledge  or the 3D structure of target/protein, which are difficult to obtain. Different from the classic methods, the proposed framework is able to automatically extract features from the data, and requires neither expert knowledge nor 3D structure of the target/protein. These salient features make the proposed framework applicable to large scale affinity data which is becoming available.

Owing to the great success of deep learning, much attention has been devoted to applying deep learning techniques for drug-target prediction. Most of the existing methods are based on topological similarity. For example, in \cite{wen2017deep} they developed a Deep Belief Network (DBN) model constructed by stacking Restricted Boltzmann Machines (RBMs). Instead of using DBN, a nonlinear end-to-end learning model named NeoDTI \cite{wan2018neodti} was proposed. NeoDTI integrates variety of information from heterogeneous network data and uses topology-preserving based representations of drugs and targets to facilitate drug-target prediction.
With the increasing popularity of graph neural networks (GNNs), researchers are adopting GNNs model for drug prediction. For example, graph convolutional network was used to model molecule based on the extraction of their circular fingerprint \cite{duvenaud2015convolutional}. By learning from molecular structures and protein sequences, Gao et al. \cite{gao2018interpretable} proposed a neural model for drug-target prediction and used a two-way attention mechanism to provide biological interpretation of the prediction. Ma et al. \cite{ma2018drug} proposed to use multi-view graph auto-encoders to obtain better inter-pretability and they also added attentive mechanism to determine the weights for each view, according to the corresponding tasks and features. Moreover, Zitnik et al. \cite{zitnik2018modeling} presented a Decagon model used for modeling polypharmacy side effects, their model constructs a multimodal graph of various interactions (i.e., protein-protein interactions, drug-target interactions) and the polypharmacy side effects. 

Among the research on deep learning for drug discovery, DeepDTA \cite{ozturk2018deepdta} and DeepCPI \cite{tsubaki2018compound} are the most relevant to our work. Both of them addressed the problem of drug-target prediction. DeepDTA takes label/one-hot encodings of compound/protein sequences as input, and trains two CNNs for the drug and target, respectively, to predict the binding affinity value of drug-target pairs. DeepCPI is originally specialized for DTI prediction, it uses a traditional GNN based on representation of $r$-radius fingerprints to encode the molecular structure of drugs, and a CNN to encode protein sequences. Attention mechanism is adopted to concatenate drug and protein representations for prediction. Here, we take measures to revise it to be used for DTA prediction.
Compared with DeepDTA and DeepCPI, the proposed framework considers both local chemical context and the topological information of drugs at the same time to improve the binding affinity prediction, by using Smi2Vec and Prot2Vec to encode the atoms in drugs and amino acids in targets, while the existing methods consider only one of these important factors. Moreover, our work also suggests a new molecular structure modeling approach that works well under our framework.


\section{Conclusions}\label{sec:conclusion}
Accurately predicting DTA is a vital and challenging task in drug discovery. In this paper, we have proposed an end-to-end deep learning framework named DeepGS for DTA prediction. It combines a GAT model to extract the topological information of molecular graph and a BiGRU model to obtain the local chemical context of drug. To assist the operations on the symbolic data, we used advanced embedding techniques (i.e., {Smi2Vec} and {Prot2Vec}) to encode the amino acids and SMILES sequences to a distributed representation. We have conducted extensive experiments to compare our proposed method with state-of-the-art models. The experimental results demonstrate that the promising performance of our proposed method. This study opens several future research directions: 1) investigating whether our method can be further improved by integrating other state-of-the-art techniques, for example, Generative Adversarial Networks; 2) extending our method to other types of problems in data mining and bioinformatics fields.

\section*{Acknowledgements}
We thank the anonymous reviewers very much for their effort in evaluating our paper. This work was supported in part by the National Key R\&D Program of China (2018YFB0204302), in part by the National Natural Science Foundation of China (No. 61972425, U1811264), and the China Scholarships Council (No. 201906130128).


\bibliography{34_paper}
\end{document}